# Predictions of photophysical properties of phosphorescent platinum(II) complexes based on ensemble machine learning approach

*Shuai Wang[1], ChiYung Yam[2,3,#], Shuguang Chen[1,2], Lihong Hu[4], Liping Li[2], Faan-Fung Hung[1,2], Jiaqi Fan[2], Chi-Ming Che[1,2,&] and GuanHua Chen[1,2]\**

[1]Department of Chemistry, The University of Hong Kong, Pokfulam, Hong Kong SAR, China

[2]Hong Kong Quantum AI Lab Limited, Pak Shek Kok, Hong Kong SAR, China

[3]Shenzhen Institute for Advanced Study, University of Electronic Science and Technology of China, Shenzhen, 518000, China

[4]School of Information Science and Technology, Northeast Normal University, Changchun, 130117, China

\*Email: ghc.hku@gmail.com (G.H.C.)

#Email: yamcy@uestc.edu.cn (C.Y.Y)

&Email: cmche@hku.hk (C.M.C.)

# Abstract

Phosphorescent metal complexes have been under intense investigations as emissive dopants for energy efficient organic light emitting diodes (OLEDs). Among them, cyclometalated Pt(II) complexes are widespread triplet emitters with color-tunable emissions. To render their practical applications as OLED emitters, it is in great need to develop Pt(II) complexes with high radiative decay rate constant ($k_r$) and photoluminescence (PL) quantum yield. Thus, an efficient and accurate prediction tool is highly desirable. Here, we develop a general protocol for accurate predictions of emission wavelength, radiative decay rate constant, and PL quantum yield for phosphorescent Pt(II) emitters based on the combination of first-principles quantum mechanical method, machine learning (ML) and experimental calibration. A new dataset concerning phosphorescent Pt(II) emitters is constructed, with more than two hundred samples collected from the literature. Features containing pertinent electronic properties of the complexes are chosen. Our results demonstrate that ensemble learning models combined with stacking-based approaches exhibit the best performance, where the values of squared correlation coefficients ($R^2$), mean absolute error (MAE), and root mean square error (RMSE) are 0.96, 7.21 nm and 13.00 nm for emission wavelength prediction, and 0.81, 0.11 and 0.15 for PL quantum yield prediction. For radiative decay rate constant ($k_r$), the obtained value of $R^2$ is 0.67 while MAE and RMSE are 0.21 and 0.25 (both in log scale), respectively. The accuracy of the protocol is further confirmed using 24 recently reported Pt(II) complexes, which demonstrates its reliability for a broad palette of Pt(II) emitters.

We expect this protocol will become a valuable tool, accelerating the rational design of novel OLED materials with desired properties.

# Introduction

Organic light-emitting diodes (OLEDs) are sustainable light sources emergingly used in displays and many other fields.[1] While the first-generation fluorescence-based OLEDs are limited to 25% internal quantum efficiency (IQE) as the ratio of singlet and triplet excitons is in 1:3 according to spin statistics,[2] one can overcome this limit by utilizing triplet excitons via phosphorescent heavy metal-based emitters. Heavy metal atoms such as iridium and platinum can induce strong spin-orbit coupling to facilitate the intersystem crossing process from the singlet to triplet excited states, and to promote radiative deactivation from the triplet excited state through phosphorescence.[3] Thus, the second-generation phosphorescent OLEDs (PhOLEDs) based on iridium/platinum emitters can achieve IQE up to 100%. Over the last decade, the research on platinum-based PhOLEDs has steadily increased in popularity and the device performances have been improved with the introduction of tetradentate cyclometalating ligands.[4-8] In fact, however, it would require substantial cost and efforts to develop high-performance PhOLED emitters experimentally, as the relationship between molecular structures and photophysical properties with interest is complicated and has not been elucidated adequately.

Theoretically, density functional theory[9] (DFT) and time-dependent density-functional theory[10] (TDDFT) are the widely-used tools to predict material properties. It enables the simulations of photophysical properties of phosphorescent emitters with reasonable balance between accuracy and efficiency. However, the results obtained from DFT or TDDFT calculations on phosphorescent emitters are still not accurate enough compared to the experimental ones.[11] To tackle this issue, machine learning (ML) algorithms, which map the complex relationship between properties and structures, could be applied.[12-14] Simplified molecular-input line-entry system (SMILES)[15] is a common tool to represent each molecule as a one-to-one string. However, it is not applicable for organometallic complexes due to the metal-ligand coordination bonds. Hence, common fingerprints/descriptors obtained via SMILES are no longer available as usual, and this poses difficulties in developing ML model for organometallic complexes. Instead, the structural parameters together with the properties based on the first-principles results are employed as features or descriptors in this work. We demonstrate that the obtained ML models improve the accuracy of predictions and thus provide an efficient tool to design novel OLED materials.[16]

Over the last decade, there were studies applying ML to predict OLED-related properties via SMILES. Alán *et al.*[17] used linear regression method to predict emission energies based on absorption energies for thermally activated delayed fluorescence (TADF) molecules. Woon *et al.*[18] established a random forest model for efficiency predictions of blue OLEDs. Lu *et al.*[19] set up light gradient boosting machine (LightGBM) models for glass transition temperature ($T_g$) and decomposition temperature ($T_d$) prediction for pure organic OLED materials. In this work, we choose features based on first-principles simulations to set up ML models. In particular, ensemble learning algorithms such as random forest (RF), adaptive boosting (Adaboost), light gradient boosting machine (LightGBM), and extreme gradient boosting (XGB) are considered and compared with conventional ML algorithms.

To further improve our ML models, stacking-based techniques are employed to enhance the generality of ML models. Finally, the prediction performances of our models were further verified by comparing to recently reported experimental data, which demonstrates reasonably good agreement and thus confirms the robustness of the models. We expect that this protocol will be beneficial for the evaluation and discovery of new PhOLED emitters efficiently and accurately.

## Methodologies

### Dataset construction and division

In this work, we mainly focus on cyclometalated Pt(II) complexes with tridentate or tetradentate ligands. Photophysical data of 206 phosphorescent Pt complexes reported in the literature are collected, including emission wavelength, photoluminescence (PL) quantum yield, and radiative decay rate constant $k_r$. They are mostly measured in degassed solutions[4] under ambient conditions and their photophysical properties distributions are shown in Fig. 1 and Section 1 of Supporting Information (SI). For molecules with structured emission band and multiple emission maxima[20], we select the lowest emission wavelength as it represents the largest emission energy. The details of data preprocessing can be found in Section 2 of SI. The emission wavelengths span a range from 430 to 661 nm, with a mean value of 526 nm. Fig. 1 shows that the values of the reported compounds distribute evenly. Similarly, the PL quantum yields of the compounds distribute evenly with a mean value of 0.421. On the other hand, $k_r$ values span over three orders of magnitude. The majority of $k_r$ values are around $1.00\times10^5$ s$^{-1}$ with a mean value of $1.20\times10^5$ s$^{-1}$. This poses a great difficulty in constructing ML models for $k_r$ prediction. To remedy this situation, care has to be taken in the data division. To maintain a balance between training and testing sets within the limited data, an improved Kennard Stone algorithm,[21] which partitions the sample set based on maximum-minimum X-Y distance (SPXY),[22] is adopted,

$$d_{xy}(p,q) = \frac{d_x(p,q)}{max_{p,q \in [1,N]} d_x(p,q)} + \frac{d_y(p,q)}{max_{p,q \in [1,N]} d_y(p,q)} \quad (p,q \in [1,N]) \quad (1)$$

where $x$ and $y$ represent the features and the target properties, respectively. $p$ and $q$ denote $p$-th and $q$-th samples in the whole dataset with $N$ samples.

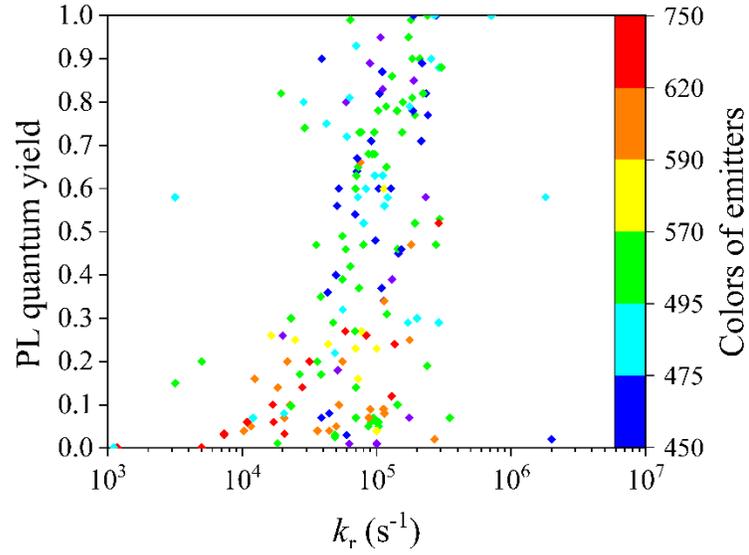

**Fig. 1. Photophysical properties distributions for PhOLED emitters**

# Features for ML

The common descriptors/fingerprints computed through SMILES are not applicable, due to their poor descriptions of coordinate bonds in organometallic complexes. Instead, in addition to structural features, first-principles simulations are employed to calculate the photophysical properties of complexes to generate molecular features in the dataset. Overall, we choose the features that are associated with the metal coordination, together with those related to the photophysical properties of these complexes, for training the ML models. As shown in Eq. 2, the rate constant is determined by the emission energy and oscillator strength of the phosphorescence transition. This is in turn highly related to the frontier orbitals of the organometallic complex. In addition, for spin-flip transition, spin-orbit couplings are essential for the transition to occur. In a Pt-emitter, the metal ion and its surrounding atoms play key roles in the phosphorescence process, thus average electron densities of these atoms are taken as features or descriptors. Besides, the coordinate bond type and coordinate bond length are considered as well. Table 1 lists all the descriptors utilized in the ML models. Details of each feature/descriptor can be found in SI Section 3.

**Table 1. Features for ML models**

| Features | Description |
|---|---|
| $\nu$ | Emission energy from the $T_1$ state to the $S_0$ state |
| coor_bond_length (N) | Coordinate bond lengths for complexes, the shortest one to longest corresponds with N from 1 to 4 |
| coor_bond_type (N) | Types of coordination (Pt-C, Pt-N, Pt-O, and Pt-Cl); The order of this series of features is correlated to coor_bond_length(N) |
| $\rho_{Pt}$ | Average electron density at Pt atom |
| $\rho\_coor$ (N) | Average electron density at the four coordination atoms |
| H_$T_1$_$S_0$ | Spin-orbit coupling constant between $T_1$ state and $S_0$ state |
| H_$T_1$_$S_1$ | Spin-orbit coupling constant between $T_1$ state and $S_1$ state |
| R_EH _excited state[a/b] | Charge-transfer descriptor interpreted in terms of the electron-hole distance in a given excitation. "a" means calculation based on literatures[23-26], "b" means calculation based on reference[27]. Small value indicates short-range excitations. |
| LAMBDA_excited state | Charge-transfer descriptor measures the spatial overlap in a given excitation.[28] Small value signifies a long-range excitation |
| CT_excited state | Charge-transfer character[23-26], 1 for completely charge-separated states; 0 for locally excited excitonic states |
| HOMO | Highest occupied molecular orbital energy |
| LUMO | Lowest unoccupied molecular orbital energy |
| $\mu$ | Molecular dipole moment |
| f | Oscillator strength of radiative transition from $T_1$ state to $S_0$ state |
| Calc_$\lambda$/$k_r$ | Calculated emission wavelength/radiative decay rate constant |
| refractive index | Refractive index to reflect the experiment testing condition |

## Machine learning algorithms

Ensemble learning strategies are employed to construct comprehensive models, including RF[29], Adaboost[30], LightGBM[31], and XGB[32]. In contrast to single weak machine learning, ensemble learning can construct faster and more accurate ML models with limited data. To assess their performance, these ensemble learning models are compared with three types of conventional ML algorithms, including support vector machines[33] (SVM), k-nearest neighbors[34] (KNN), and kernel ridge regression[35] (KRR). In addition, stacking techniques are adopted to enhance the generality of ML models.[36] After SPXY data division, 80% of the dataset was selected for model training, and the remaining 20% was used as an independent test set. 10-fold cross-validation was adopted to improve the stability of the obtained ML models. The hyperparameters are tuned by python library hyperopt[37]. Details of the hyperparameter optimization can be found in Section 4 of SI. Performances on the models were evaluated based on the squared correlation coefficient ($R^2$), the root mean square error (RMSE), and mean absolute error (MAE).

# First-principles simulations

Gaussian16 program package[38] was utilized for all geometry optimizations, and ADF2021 package[39] was employed to calculate the phosphorescence of the Pt(II) complexes. The optimized geometries of ground state ($S_0$) and excited states ($T_1$) were calculated by DFT and TDDFT, respectively, with B3LYP functional[40]. Relativistic effects were considered for Pt atom using the Stuttgart basis set[41] and pseudopotential. 6-31G* atomic basis set[42,43] was used for all other atoms. Solvation effects were taken into account using polarizable continuum model (PCM)[44]. With the optimized structures, emission properties were calculated using the ADF2021 package[39]. For phosphorescence transitions, spin-orbit coupling (SOC) was treated as a perturbation based on the scalar relativistic orbitals.[45] Triple-zeta polarized Slater-type basis set for all atoms and PBE0 functional were used in the TDDFT calculations.[46] In addition, matrix effects were considered using the COSMO continuum solvation model in relevant experimental testing medium.[47] With the emission energy and oscillator strength obtained, $k_r$ can be calculated as follows.

$$k_r = \frac{2\pi v^2 e^2}{\varepsilon_0 m c^3} f \qquad (2)$$

where $v$ is the emission energy from the lowest triplet state ($T_1$) to the ground state ($S_0$); $e$ denotes the elementary electric charge; $\varepsilon_0$ is the vacuum permittivity; $m$ represents the mass of electrons; $c$ is the speed of light, and $f$ is the oscillator strength of $T_1 \rightarrow S_0$ transition.

# Results and discussion

ML models are constructed using different algorithms, including RF[29], AdaBoost[30], LightGBM[31], XGB[32], SVM[33], KNN[34] and KRR[35]. The performance of each model regarding emission wavelength predictions after 10-fold cross-validation can be found in Table 2. Apparently, LightGBM gives the best performance in terms of both correlation coefficient and errors. Basically, LightGBM improves the efficiency and scalability of gradient boosting (GB) algorithm without sacrificing its inherited effective performance, which is suitable for rapid assessment of the dataset. See details of the performance of each ML algorithm in Section 4 of SI.

We further analyze the importance of each feature as listed in Fig. 2. It can be seen that the calculated emission wavelength shows the most remarkable contribution in determining the target property. In addition, LUMO/HOMO energy, oscillator strength and spin-orbit coupling constant contribute significantly to the model as well. Others are primarily transition-related features. For instance, the charge-transfer features[23-26] (CT series in Fig. 3) at the transitions from the third triplet state and first singlet state to the ground state both contribute to the optimal model.

**Table 2. Performance of each ML algorithm on the prediction of emission wavelength**

| ML models | Independent testing set[a] | | |
| --- | --- | --- | --- |
| | MAE (nm) | RMSE (nm) | $R^2$ |
| KNN_Uniform | 27.38±1.17 | 40.90±1.52 | 0.57±0.04 |
| KNN_Distance | 17.15±1.03 | 28.33±2.40 | 0.80±0.04 |
| SVM | 20.52±2.12 | 32.00±2.91 | 0.76±0.05 |
| KRR | 22.62±0.67 | 28.55±0.80 | 0.82±0.02 |
| RF | 7.98±0.72 | 11.81±1.15 | 0.97±0.01 |
| **LightGBM** | **5.57±0.63** | **8.73±1.24** | **0.98±0.01** |
| Adaboost | 13.28±0.47 | 16.01±0.65 | 0.95±0.01 |

[a]The standard deviations are calculated by the difference in the prediction of each fold.

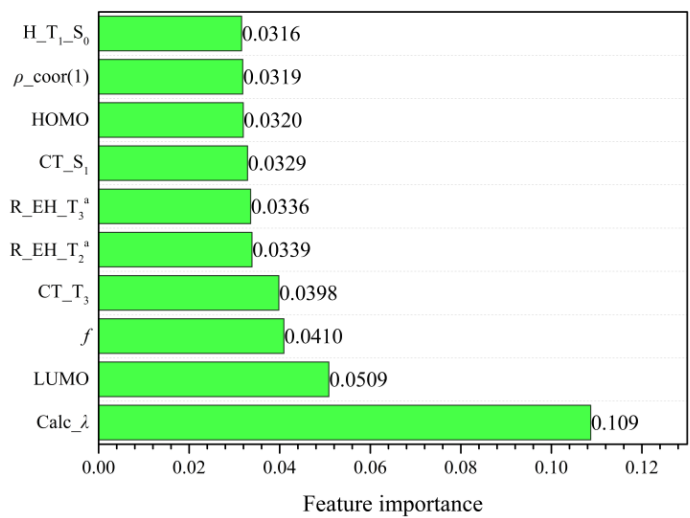

**Fig. 2. Ten most important features for emission wavelength extracted from LightGBM-based ML model**

The stacking approach with two-layer architecture is used to further improve the generality and stability of the emission wavelength model. Here, the wavelength predicted by LightGBM model is concatenated with all the features utilized in the base-learner layer. The stacking architecture is shown in Fig. 3. With the new features, four different types of meta-learners are tested and the results are listed in Table 3. SVM as meta-learner shows the highest correlation coefficient and lowest errors, and it is selected as the optimal meta-learner. Finally, Fig. 4 plots the performance of the stacking model in predicting emission wavelength on the independent testing set. It is clear that the ML predicted results agree well with experiments and show the substantial improvement over simulation results. Excellent consistency between ML predicted results and experimental results with $R^2$ of 0.96, MAE of 7.21 nm and RMSE of 13.00 nm is obtained.

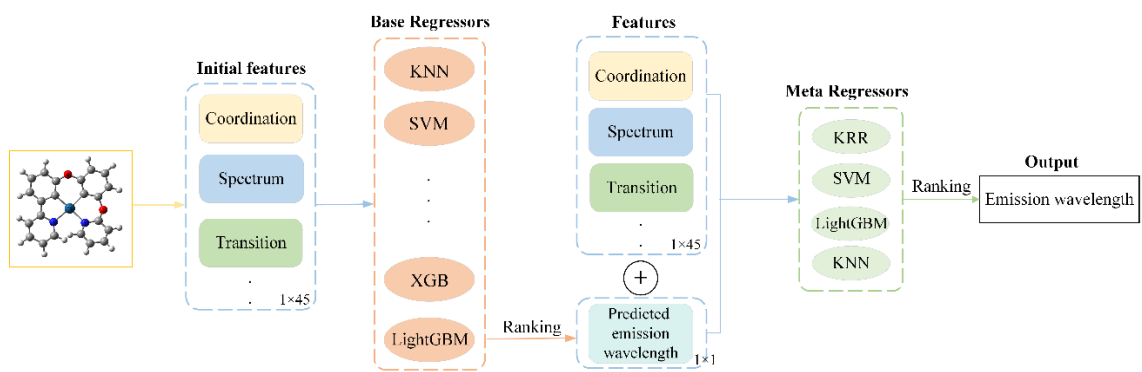

**Fig. 3. Proposed stacking architecture for emission wavelength prediction**

**Table 3. Meta-regressors comparison on emission wavelength in stacking**

| | Independent testing set[a] | | |
|---|---|---|---|
| ML models | MAE | RMSE | $R^2$ |
| KRR | 13.10±0.74 | 15.85±0.78 | 0.94±0.01 |
| **SVM** | **7.22±0.77** | **13.00±1.33** | **0.96±0.01** |
| LightGBM | 12.37±0.70 | 16.09±1.13 | 0.94±0.01 |
| KNN_Distance | 11.96±1.01 | 17.67±2.76 | 0.92±0.03 |

[a]The standard deviations are calculated by the difference in the prediction of each fold.

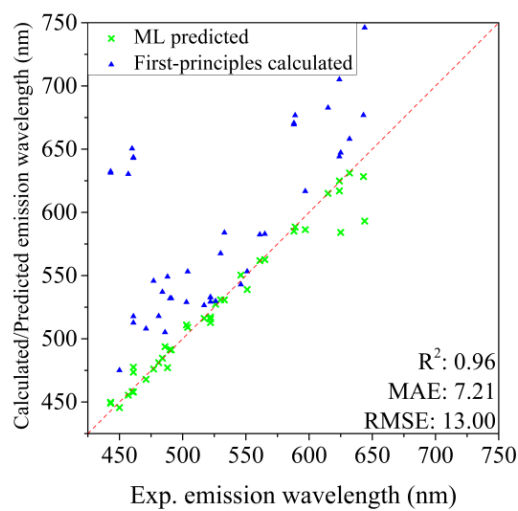

**Fig. 4. SVM performances of emission wavelength calibration model on the independent testing set**

Radiative decay rate constant $k_r$ of OLED emitter is another fundamental quantity that determines the performance of OLEDs. Emitters with large $k_r$ values are preferred to avoid bimolecular quenching processes in OLED that could lead to severe efficiency roll-off at high luminance and material degradation . To predict the radiative decay rate constant $k_r$, eight models are constructed using different ML algorithms whose performance are listed in Table 4. It is found that KNN, RF and XGB have similar correlation coefficients and errors. However, RF holds the best prediction stability when compared to the other two models. RF is usually robust and stable for the outliers and noises, a property that is suitable for the case of $k_r$ since the experimental values span over three orders of magnitude and are not balanced. Essentially, RF creates trees on the subset of the data and combines the output of all the trees, which reduces overfitting problem in decision trees and the variance. Similarly, we analyze the importance of different features as plotted in Fig. 5. Again, the emission energy has the highest contribution as it is one of the key terms determining the rate constant. On the other hand, the features about electron-hole distance[23-26] from $T_2$ and $S_1$ to ground state (R_EH_T$_2$[a], R_EH_S$_1$[a]), which describe the average electron−hole separation, have a considerable importance compared to the other features. In contrast, the calculated rate constant ranks lower, which indicates the relative inaccurate description of singlet-triplet transition in TDDFT. Besides, oscillator strength, bond lengths of coordination bonds, and average electron densities of the coordination atoms are also important features in the model.

**Table 4. Performance of each ML algorithm on the prediction of radiative decay rate constants**

| ML models | Independent testing set[a] | | |
| --- | --- | --- | --- |
| | MAE | RMSE | $R^2$ |
| KNN_Uniform | 0.28±0.01 | 0.32±0.01 | 0.23±0.05 |
| KNN_Distance | 0.19±0.01 | 0.22±0.01 | 0.74±0.06 |
| SVM | 0.22±0.01 | 0.26±0.01 | 0.71±0.08 |
| KRR | 0.22±0.01 | 0.27±0.01 | 0.57±0.04 |
| **RF** | **0.16±0.01** | **0.22±0.01** | **0.74±0.05** |
| LightGBM | 0.17±0.01 | 0.24±0.01 | 0.65±0.06 |
| Adaboost | 0.23±0.01 | 0.27±0.02 | 0.63±0.09 |
| XGB | 0.16±0.01 | 0.21±0.03 | 0.73±0.07 |

[a]The standard deviations are calculated by the difference in the prediction of each fold.

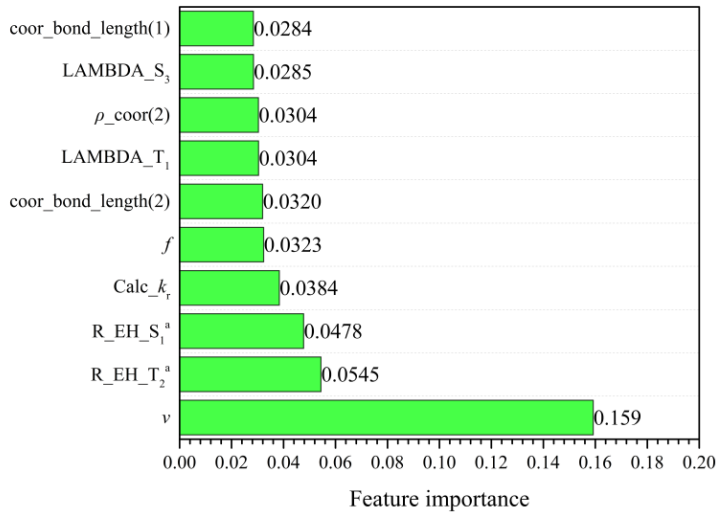

**Fig. 5. Ten most important features for radiative decay rate constants extracted from RF-based ML model**

To further improve the prediction performance, stacking technique is adopted with two layers. In practice, the benefit of stacking is that it can harness the capabilities of a range of well-performing models on a classification or regression task and make predictions that have better performance than any single model in the base-learner layer. As shown in Fig. 6, the stacking architecture includes the aggregation of AdaBoost, LightGBM, RF, and XGBoost as the base learner layer. These ensemble algorithms exhibit superior results on the independent testing set so that their predicted $k_r$ values are selected as the features for the subsequent training. This kind of selection would take advantage of different algorithms to get more stable and precise predictions. Then four different types of meta-learners are tested and compared. The results are shown in Table 5. Meta-learner KNN_Distance, among the four algorithms, exhibits the best performance on error items after stacking. The results with meta-learner KNN_Distance can be seen in Fig 7. Clearly, our ML model significantly improves the prediction of radiative decay rate constants with MAE of 0.21, RMSE of 0.25 and $R^2$ of 0.67.

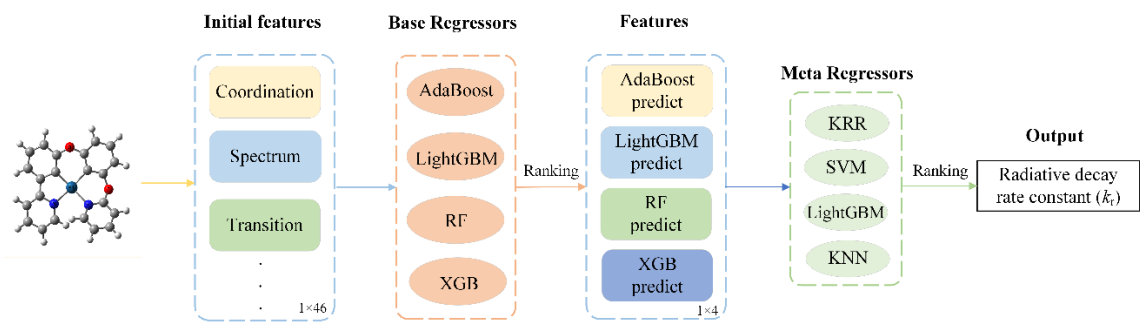

**Fig. 6. Proposed stacking architecture for $k_r$ prediction**

**Table 5. Meta-regressors comparison on radiative decay rate constants in stacking**

| | Independent testing set[a] | | |
|---|---|---|---|
| ML models | MAE | RMSE | $R^2$ |
| KRR | 0.95±0.01 | 1.26±0.01 | 0.77±0.00 |
| SVM | 0.45±0.05 | 0.51±0.05 | 0.01±0.01 |
| LightGBM | 0.22±0.01 | 0.27±0.01 | 0.53±0.06 |
| **KNN_Distance** | **0.21±0.01** | **0.25±0.01** | **0.67±0.04** |

[a]The standard deviations are calculated by the difference in the prediction of each fold.

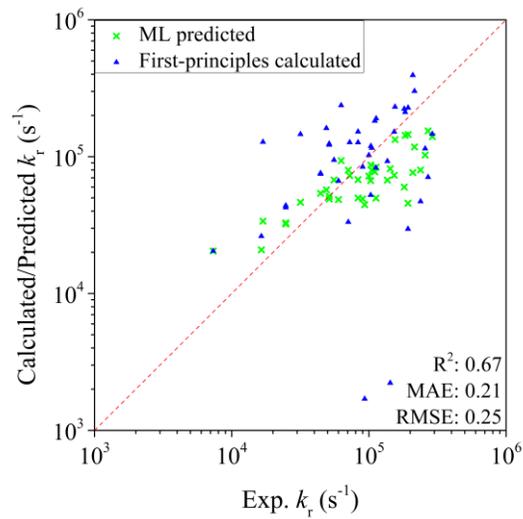

**Fig. 7. KNN performances of radiative decay constant calibration model on the independent testing set**

PL quantum yield is defined as the ratio of the number of photons emitted to the number of photons absorbed, which is a prominent property of OLED emitters. From the theoretical perspective, accurate prediction of PL quantum yield of phosphorescent Pt emitters using first-principles simulations is challenging[48]. In this work, quantum yield is therefore not chosen as one of the features. Instead, features listed in Table 1 are used, except calculated emission wavelength and $k_r$. Consequently, eight models are constructed using different ML algorithms. Table 6 illustrates the performance of each model. It is found that though LightGBM has relatively good performances in terms of errors and correlation coefficients, it is not as stable as RF. Feature importance analysis of the RF model for PL quantum yield is shown in Fig. 8. As expected, emission energy is the most important feature, followed by average electron density of coordination atoms, LUMO/HOMO energies, spin-orbit coupling constant, and other features that are closely related to the radiative transition.

**Table 6. Performance of each ML algorithm on the prediction of PL quantum yield**

| ML models | Independent testing set[a] | | |
| --- | --- | --- | --- |
| | MAE | RMSE | $R^2$ |
| KNN_Uniform | 0.21±0.01 | 0.26±0.01 | 0.37±0.03 |
| KNN_Distance | 0.16±0.00 | 0.21±0.01 | 0.59±0.03 |
| SVM | 0.21±0.00 | 0.24±0.01 | 0.63±0.04 |
| KRR | 0.19±0.00 | 0.23±0.00 | 0.57±0.03 |
| **RF** | **0.11±0.00** | **0.15±0.01** | **0.81±0.02** |
| LightGBM | 0.09±0.01 | 0.13±0.01 | 0.83±0.04 |
| Adaboost | 0.13±0.00 | 0.17±0.01 | 0.74±0.02 |
| XGB | 0.13±0.01 | 0.18±0.01 | 0.74±0.04 |

[a]The standard deviations are calculated by the difference in the prediction of each fold

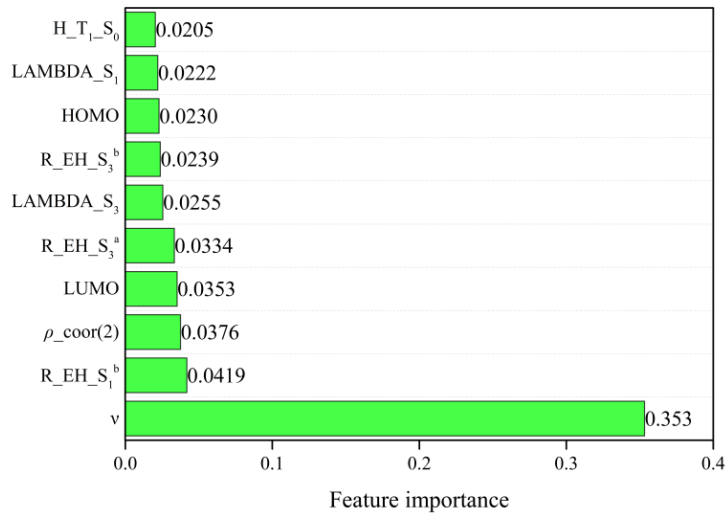

**Fig. 8. Ten most important features for PL quantum yield extracted from RF-based ML model**

Next, similar to the emission wavelength, the stacking approach with two-layer architecture is used to further improve the prediction capability of the PL quantum yield model. Here, the quantum yield predicted by RF model is concatenated with all other features utilized in the base-learner layer. The stacking architecture is shown in Fig. 9. With the new features, different types of meta-learners are tested and the results are listed in Table 7. RF as meta-learner shows the highest correlation coefficient and lowest errors, and it is selected as the optimal meta-learner. Finally, Fig. 10 plots the performance of the stacking model in predicting PL quantum yield on the independent testing set. Satisfactory consistency between ML predicted results and experimental measurements is achieved with $R^2$ of 0.81, MAE of 0.11 and RMSE of 0.15.

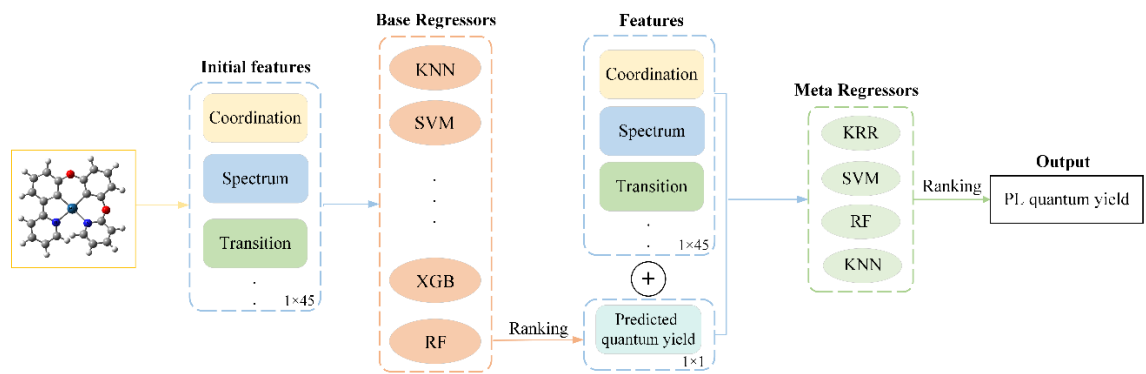

**Fig. 9. Stacking architecture for PL quantum yield prediction**

**Table 7. Meta-regressors comparison on quantum yield in stacking**

| ML models | Independent testing set[a] | | |
|---|---|---|---|
| | MAE | RMSE | $R^2$ |
| KRR | 0.13±0.01 | 0.17±0.00 | 0.78±0.01 |
| SVM | 0.15±0.01 | 0.17±0.01 | 0.81±0.04 |
| **RF** | **0.11±0.00** | **0.15±0.01** | **0.81±0.02** |
| KNN_Distance | 0.15±0.01 | 0.19±0.01 | 0.65±0.04 |

[a]The standard deviations are calculated by the difference in the prediction of each fold

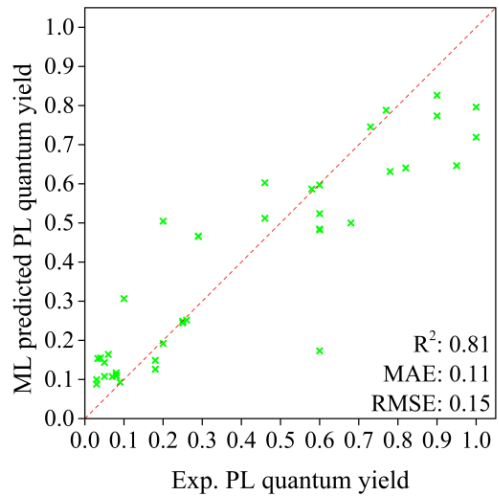

**Fig. 10. RF performance of PL quantum yield prediction model on the independent testing set**

Realizing accurate photophysical properties predictions of new complexes prior to experimental synthesis will be of great importance for the development of OLED materials. To assess the performance of our model, twenty-four recently reported complexes are collected as external samples to evaluate the generalization capacity of the three vital photophysical properties. More detailed information on these Pt-based complexes can be found in Section 5 of SI. Fig. 11 (a)-(c) shows respectively the prediction of the three photophysical properties. It can be seen that the ML models satisfactorily predict the phosphorescence properties of these latest studied Pt-based complexes over diverse scaffolds and therefore confirms the generalization abilities of our models.

For the emission wavelength predictions, $R^2$ of 0.81, RMSE of 16.58 nm and MAE of 12.94 nm are achieved. When compared with the mean value of 526 nm, the result is accurate enough for evaluation of the performance and screening of these emitters. On the other hand, regarding radiative decay rate constant $k_r$, MAE of 0.21 and RMSE of 0.24 (both in log scale) are very similar to those in the testing set and acceptable for $k_r$ predictions. For the PL quantum yield, an outstanding performance with MAE of 0.12 and RMSE of 0.15 is realized, except that there are four outliers as highlighted in Fig. 10 (c). Prediction details of the external samples can be seen in Table S10 of SI. Overall, the developed ML models in this work not only give exceptional performance on the independent testing set, but also demonstrate satisfactory results on the external testing set.

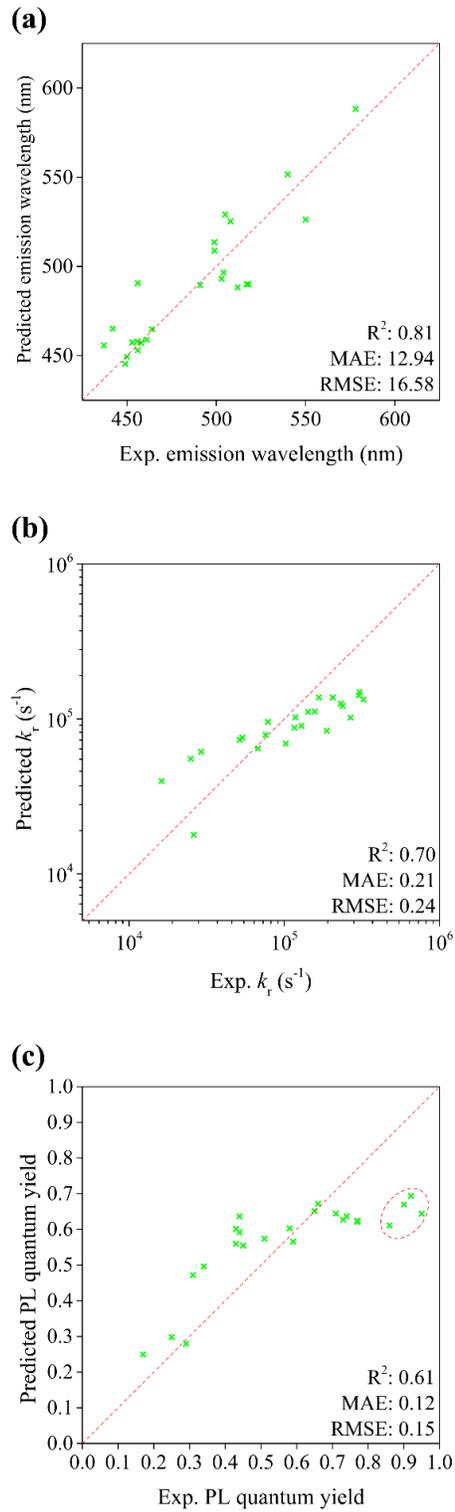

**Fig. 11. Performances of optimal ML models for (a) emission wavelength, (b) radiative decay rate constant, and (c) PL quantum yield on external samples**

# Conclusion

In summary, a general protocol of first-principles calculations and machine learning models is constructed to predict the photophysical properties of phosphorescent Pt(II) emitters. Ensemble ML models (XGBoost, LightGBM, RF, and AdaBoost) are utilized based on a dataset of 206 Pt-based emitters. These are compared with conventional ML models (SVM, KNN, and KRR) to demonstrate their performance. The feature analysis reveals that emission energy, coordination, and spin-orbit coupling constant show significant contributions in the prediction. To further improve the performance, the stacking methods are implemented. Finally, recently reported Pt-complexes are employed as external samples to evaluate the generalization capability of the ML models, which indicates the robustness of the protocol. This work presents the first ML protocol for predicting and evaluating three important photophysical properties of Pt-emitters by employing ensemble ML algorithms. We expect the protocol would be beneficial to scientists in designing novel Pt-emitters with superior performances and thus help discover novel OLED materials.

## Data and Software Availability

Gaussian 16 package[38] can be found from http://gaussian.com and ADF2021 package[39] (v.2021.102) is a commercial software, for which a free trial can be requested at http://www.scm.com.

## Acknowledgement

Financial support from the RGC General Research Fund under grant no. 17309620 and Hong Kong Quantum AI Lab Ltd. is gratefully acknowledged.

# For Table of Contents Use Only

**Predictions of photophysical properties of phosphorescent platinum(II) complexes based on ensemble machine learning approach**

Shuai Wang, ChiYung Yam*, Shuguang Chen, Lihong Hu, Liping Li, Faan-Fung Hung, Jiaqi Fan, Chi-Ming Che* and GuanHua Chen*

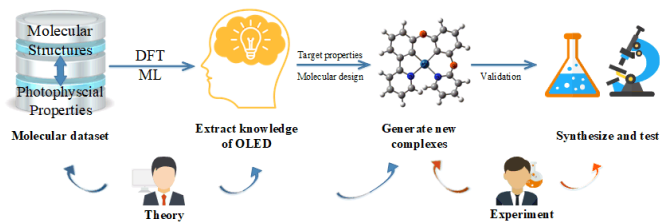